%% file: Run_This_Example.tex
\pdfoutput=1
\documentclass[Alon2,ChapterTOCs,singlecolor,11pt]{Alon}
\usepackage{fixltx2e,fix-cm}
\usepackage[english]{babel}
\usepackage{epigrafica}
\usepackage[LGR,OT1]{fontenc}
\usepackage{mathptmx,txfonts}
\usepackage{lmodern}
\usepackage{amssymb}
\usepackage{txfonts} % or any other font package
\usepackage{graphicx}
\usepackage{subfigure}
\usepackage{makeidx}
\usepackage{multicol}
\usepackage{booktabs}
\usepackage{multirow}
\usepackage{enumitem}
\usepackage{comment}

\usepackage{cite}

\usepackage[table,xcdraw]{xcolor}

\usepackage{textcomp}
\usepackage{comment}
\usepackage{enumerate}
\usepackage{caption}

\usepackage{rotating}

\usepackage{changepage}% for text width to include margin

\makeindex

\usepackage{version}

\usepackage{lipsum} % To generate some text for the article via \lipsum

\usepackage{tikz,xcolor}
\usepackage{url} 

\definecolor{lime}{HTML}{A6CE39}
\DeclareRobustCommand{\orcidicon}{%
	\begin{tikzpicture}
	\draw[lime, fill=lime] (0,0) 
	circle [radius=0.16] 
	node[white] {{\fontfamily{qag}\selectfont \tiny ID}};
	\draw[white, fill=white] (-0.0625,0.095) 
	circle [radius=0.007];
	\end{tikzpicture}
	\hspace{-2mm}
}

\foreach \x in {A, ..., Z}{%
	\expandafter\xdef\csname orcid\x\endcsname{\noexpand\href{https://orcid.org/\csname orcidauthor\x\endcsname}{\noexpand\orcidicon}}
}

\usepackage{epstopdf}% To incorporate .eps illustrations using PDFLaTeX, etc.
\usepackage[caption=false]{subfig}% Support for small, `sub' figures and tables
\renewcommand\bibfont{\fontsize{10}{12}\selectfont}% To set the list of references in 10 point font using natbib.sty

\title{Taylor and Francis Book Chapter}
\begin{document}

% \frontmatter

\begingroup
\pagestyle{empty}
\cleardoublepage
\endgroup

\maketitle %Computer Vision and Image Analysis for Industry 4.0

%%%Placeholder for front matter

%\halftitle

%\booktitle Computer Vision and Image Analysis for Industry 4.0

%\locpage

\include{frontmatter/dedication}
%\cleardoublepage
\setcounter{page}{1} %previous pages reserved for frontmatter to be added later
\tableofcontents
%\listoffigures
%\listoftables
%\include{frontmatter/foreword}
%\include{frontmatter/preface}
%\include{frontmatter/contributor}
%\include{frontmatter/authorbio}

% \mainmatter

% Clear Blank Page

\let\cleardoublepage\clearpage

%\part{This is a Part}

\include{chapter1/ch1}

%\setcounter{chapter}{10}
%\include{chapter2/ch2}

% \bibliographystyle{plain}
% \bibliography{Run_This_Example.bib}

%\printindex
%\cleardoublepage
\end{document}

%% file: chapter1/ch1.tex
% new add
% \newcommand*\rot{\rotatebox{90}}
% \newcommand*\OK{\ding{51}}

% \renewcommand*\rot{\rotatebox{90}}
% \renewcommand*\OK{\ding{51}}

\chapterauthor{Rajana Akter*}{American International University, Dhaka, Bangladesh\\
Email: rejana.akter@gmail.com, *Corresponding Author}
\chapterauthor{Shahnure Rabib}{American International University, Dhaka, Bangladesh\\
Email: shahnoorrabib89@gmail.com}
\chapterauthor{Rahul Deb Mohalder}{Khulna University, Khulna, Bangladesh\\
Email: rahul@ku.ac.bd}
\chapterauthor{Laboni Paul}{Khulna University, Khulna, Bangladesh\\
Email: laboni1124@cseku.ac.bd}
\chapterauthor{Ferdous Bin Ali}{visie.tech, Dhaka, Bangladesh\\
Email: hridoyferdous@yahoo.com}

\chapter{SCGNet-Stacked Convolution with Gated Recurrent Unit Network for Cyber Network Intrusion Detection and Intrusion Type Classification}

\chaptermark{SCGNet Intrusion Detection and Intrusion Type Classification}

\chapterinitial {I}ntrusion detection system (IDS) is a piece of hardware or software that looks for malicious activity or policy violations in a network. It looks for malicious activity or security flaws on a network or system. IDS protects hosts or networks by looking for indications of known attacks or deviations from normal behavior (Network-based intrusion detection system, or NIDS for short). Due to the rapidly increasing amount of network data, traditional intrusion detection systems (IDSs) are far from being able to quickly and efficiently identify complex and varied network attacks, especially those linked to low-frequency attacks. The SCGNet (Stacked Convolution with Gated Recurrent Unit Network) is a novel deep learning architecture that we propose in this study. It exhibits promising results on the NSL-KDD dataset in both task, network attack detection, and attack type classification with 99.76\% and 98.92\% accuracy, respectively. We have also introduced a general data preprocessing pipeline that is easily applicable to other similar datasets. We have also experimented with conventional machine-learning techniques to evaluate the performance of the data processing pipeline.

\section{Introduction}
Network security is a dynamic field where new attack types constantly emerge and need to be countered. The data set contains four main categories of attacks: DoS, Probe, User to Root (U2R), and Remote to Local (R2L). To block traffic travelling to and from the target system is the aim of a DoS attack. The IDS must shut down in order to defend itself due to the unusually large volume of traffic. This prevents a network from being accessed by normal traffic. The network may get overloaded and shut down when an online store gets a lot of orders on a day when there is a significant discount, prohibiting paying customers from making any purchases. This assault happens the most frequently in the data set.

On the other hand, an attempt to obtain data from a network is referred to as a probe or surveillance attack. The goal is to appear as a robber and steal important information, whether it be financial information or customer personal information. On a similar note, In a U2R attack, a regular user account is used to try root access to the system or network. The attacker tries to take advantage of a system's flaws in order to gain access or root privileges. The goal of an R2L attack is to get physical access to a remote machine. Even if they do not have local access to the system or network, an attacker tries to "hack" their way in.

This has led to the creation of software called Network Intrusion Detection Systems (NIDS), which helps identify security breaches from traﬀic packets in real-time. IDS uses machine learning shallow learning (single feed-forward networks), which is dependent on feature engineering to get the features for the ML model \cite{10_vasilomanolakis2015taxonomy}. IDS uses feature selection, classification methods, decision trees, SVM, K-nearest, and shallow learning (single feed-forward networks). Deep learning’s superior representational capabilities have received a lot of attention recently. Deep learning has produced outstanding results in many fields, including image identification and natural language processing (NLP) \cite{2.2_huang2017densely}.

However, network traﬀic data is often one-dimensional, making it diﬀicult to use convolutional neural networks (CNN) to detect cyberattacks \cite{12_naseer2018enhanced, 14_li2020building}. To address this, research has focused on the KDD CUP 99 dataset, which is a well-liked standard for classifier precision, but has a number of flaws that make it unsuitable for usage in contemporary settings. Tavallaee et al. has developed NSL-KDD, a more balanced resampling of KDD-99 that places emphasis on cases that are likely to be overlooked by classification algorithms trained on the basic dataset \cite{6_tavallaee2009detailed}.

\subsection{Our Contribution:}
In our research, we have developed a data pipeline alongside a novel deep-learning model that can be used for defined attack detection with great accuracy. Our contribution goes as follows: 

\begin{itemize}
  \item We have proposed a data pipeline that can handle missing value, class imbalance and other preprocessing steps for any kind of data.
  \item We have created a revolutionary architecture referred to as SCGNet (Stacked Convolution and GRU Network) that achieves state-of-the-art results for both binary and multiclass classification.
  \item We have defined the network using Random Search \cite{1_omalley2019kerastuner} for architecture definement Hyperband \cite{2_li2017hyperband} for hyperparameter search.
  \item We have also experimented with traditional machine learning classifiers to classify different sentiments.
\end{itemize}

We organized this paper as follows. Section \ref{sec_literature_survey} about previous methods and works. In the section \ref{sec_methodology} our proposed methodology. Section \ref{sec_result_analysis} we analyzed our outcomes and contrasted them with other researchers' work. Finally, in section \ref{sec_conclusion_and_future_work}, we gave the summary and plan of this research.

\section{Literature Survey}
\label{sec_literature_survey}
Intrusion Detection system is a vital research area that combined with Machine Learning, Deep Learning and Data mining approaches may be used to improve prediction performance and also detect the type of attack. As earlier classification algorithms used to exact features and deep learning methodologies demonstrate their effectiveness. Anomaly detection is accomplished using a variety of machine-learning methods. Many researchers have Concentrated on deep learning methods to build effective IDSs. identifying doubtful network activity is the aim of attack detection. Several machine learning methods are used as the most popular intrusion prevention strategy to lower the mistake rate \cite{1.1_chung2015heuristic}. 

Ma T; Cheng J; Classified the type of attack using a Deep Learning Neural Network and an extractor of features using spectral clustering. The DNN Network outperformed the SVM, BPNN, and RF Network in terms of accuracy \cite{3_laghrissi2021intrusion}. However Sydney; proposes a wireless IDS system based on a Feed-Forward Deep Neural Network (FFDNN) and compares it to standard machine learning algorithms such as Random Forest(RF), Support Vector Machine (SVM), Naive Bayas(NB), Decision Tree (DT) and K-nearest Neighbor (KNN). Binary multiclass attacks are included in the experimental studies. The solution performed well on the UNSW-NB15 and AWID datasets with 87.10\% and 77.16\% accuracy for binary and multiclass classification respectively \cite{5_kunang2021attack}. 

% On the other hand, Chickerbene proposed two models for intrusion detection and classification systems, which reduce the number of features in the input data based on a new feature selection algorithm and TIDES-A which is a dynamic algorithm to compute the exact time for nodes cleaning states and restricts the exposure window of the nodes \cite{15_chkirbene2020tidcs}.

Another approach was based on the dataset KDD99, Fatima Ezzahra; proposed three models: LSTM, LSTM-PCA, and LSTM-MI. These methods were put to the test for the categorization of objects into binary and multiclass categories, and the results showed that PCA-based models generated the best results. PCA- based architectures, particularly those with two components, produced the best results and multiclass classification with 99.44\% and 99.39\% respectively \cite{7_divekar2018benchmarking}.

% Based on the Deep Learning method LSTM, Alaeddine developed a brand-new idea for NIDS that will recognize attacks and keep a long-term memory of them in order to block more attacks while also treating each sort of attack differently \cite{16_boukhalfa2020lstm}.

Additionally, Abhishek examined the UNSW-NB15 dataset, which has 10 contemporary attack classes and a less skewed distribution of targets. To make UNSW-NB15 adoption in further studies easier, F1 performance has been compared to models typically trained on KDD-99 and NSL-KDD. The results of this research provide enough room for performance to be optimized by employing alternative strategies across the machine learning pipeline, which overcomes the underwhelming performance of classifiers trained on NSL-KDD and UNSW-NB15 \cite{11_sukumar2018network}. 

However, Yasi's proposed IDS model, which uses DL and a DAE for the pre-training process and fine-tuning utilizing DNN through the process of HPO, improves the results of attack classification in intrusion detection. Additionally, the ideal feature extraction technique for developing a successful DL IDS is also taken into account \cite{12_naseer2018enhanced}. The results of all methods are in Table. \ref{tab:nsid}.

\begin{center}
\begin{table}[ht]  
\caption{Network Security Intrusion Detection Literature Survey Result} % title name of the table  
\label{tab:nsid}
\centering % centering table  
\begin{tabular}{l c c rrrrrrr} % creating 10 columns  
% \begin{tabular}{m{4em} m{3cm} m{3cm} m{1.5cm} m{1.5cm} m{1.5cm} m{1.5cm}}
\hline\hline   
\rotatebox{90}{Reference} & \rotatebox{90}{Dataset} & \rotatebox{90}{Algorithm} & \rotatebox{90}{Accuracy(\%)} & \rotatebox{90}{Precision} & \rotatebox{90}{Recall} & \rotatebox{90}{F-1 score}
\\ [0.9ex]  
\hline \hline  

% Entering row  
 & & \begin{tabular}[c]{@{}c@{}}LSTM-PCA\\ (Binary)\end{tabular}& 99.44 & - & - & 0.99 \\[-0.3ex]  
\raisebox{1.5ex}{\cite{7_divekar2018benchmarking}} & \raisebox{1.5ex}{KDD-99
}& \begin{tabular}[c]{@{}c@{}}LSTM-PCA\\ (Multi)\end{tabular} & 99.39 & - & - & 0.99 \\[1ex] 
\hline  
% Entering  row  
 & & \begin{tabular}[c]{@{}c@{}}k-means\\ Algorithm\end{tabular} & 53.271028 & 0.005566 & 0.004673 & - \\[-0.3ex]  
\raisebox{1.5ex}{\cite{2.2_huang2017densely}} & \raisebox{1.5ex}{KDD-99}& \begin{tabular}[c]{@{}c@{}}IGKM\\ Algorithm\end{tabular}  
& 72.913043 & 0.352941 & 0.268869 & - \\[1ex] 
\hline  
% Entering row  
 & \begin{tabular}[c]{@{}c@{}}UNSW-NB15\\ (train-Bina)\end{tabular} &  & 94.34 & - & - & - \\[-0.5ex]  
\raisebox{1ex}{\cite{5_kunang2021attack}} & \begin{tabular}[c]{@{}c@{}}UNSW-NB15\\ (test-Bina)\end{tabular} & \raisebox{1ex}{FFDNN} & 87.48 & - & - & - \\[-.1ex]  
 & \begin{tabular}[c]{@{}c@{}}UNSW-NB15\\ (train-Multi)\end{tabular} &  & 80.80 & - & - & - \\[-.1ex]  
 & \begin{tabular}[c]{@{}c@{}}UNSW-NB15\\ (test-Multi)\end{tabular} &  & 75.83 & - & - & - \\[1ex] 
\hline  
% Entering row  
 & & RF & 99.72 & 88.67 & - & - \\[-0.3ex]  
\raisebox{1.5ex}{\cite{3_laghrissi2021intrusion}} & \raisebox{1.5ex}{NSL-KDD}& \begin{tabular}[c]{@{}c@{}}SCDNN\\ Algorithm\end{tabular} & 97.23 & 52.66 & - & - \\[1ex] 
\hline  
 % Entering row  
 & KDDCUP-99 & & 99.7 & - & - & - \\[-1ex]  
\raisebox{1.5ex}{\cite{9_ma2016hybrid}} & UNSWNB-15 & \raisebox{1.5ex}{LSTM} & 99.7 & - & - & - \\[1ex] 
\hline  
% Entering row  
 & NSL-KDD (test) & & 85.797 & 88.0 & - & - \\[-1ex]  
\raisebox{1.5ex}{\cite{4_ma2017learning}} & UNSW-NB15(test) & \raisebox{1.5ex}{\begin{tabular}[c]{@{}c@{}}Two-Stage\\ Ensemble\end{tabular}} & 91.27 & 91.60 & - & - \\[1ex] 
\hline  
% Entering row  
 & NSL-KDD & & 83.33 & 86.02 & 82.32 & 82.04 \\[-1ex]  
\raisebox{1.5ex}{\cite{12_naseer2018enhanced}} & CSE-CIC-IDS2018 & \raisebox{1.5ex}{DNN} & 95.79 & 95.38 & 95.79 & 95.11 \\[1ex]  
\hline  
 % Entering row  
 & NSL-KDD & & 98 & 94 & - & - \\[-1ex]  
\raisebox{1.5ex}{\cite{6_tavallaee2009detailed}} & UNSW-NB15 & \raisebox{1.5ex}{TIDCS} & 91 & - & - & - \\[1ex] 

\hline  
 % Entering row  
\cite{10_vasilomanolakis2015taxonomy} & NSL-KDD & LSTM & 99.999 & - & 99.973 & 99.986 \\[1ex]

% [1ex] adds vertical space  
\hline % inserts single-line 
\hline  
\end{tabular}  
\end{table}  
\end{center}

\section{Methodology}
\label{sec_methodology}

\begin{figure}
	\centering
	\includegraphics[ width=0.95\textwidth, height=0.70\textwidth]{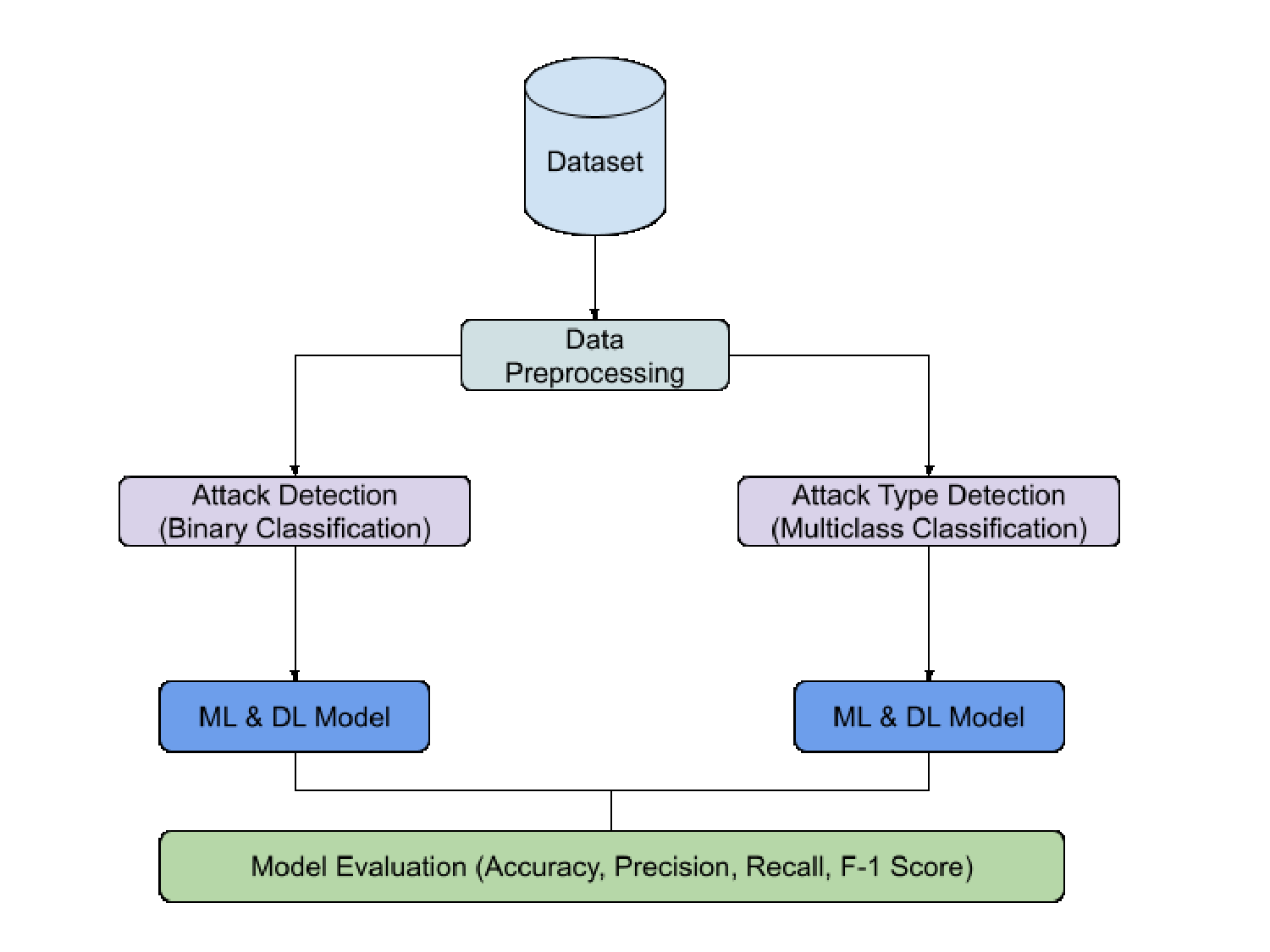}
	\caption{Our Proposed Workflow Diagram.}
	\label{fig:workflow}
\end{figure}

In our paper, we proposed different types of properties to train our model in order to find Network attack types detection performance. We have deployed machine learning models and deep learning models such as SCGNet, AdaBoost, CNN, and XGBoost to detect Network attack types and address imbalance dataset issues. In  Fig. \ref{fig:workflow} we illustrated our proposed workflow diagram. 

\subsection{Dataset Description}
There have been several datasets of this kind before the NSL-KDD one. The KDD Cup was an international contest for data mining and knowledge discovery tools. In order to gather traffic data, this competition was started in 1999. The competition's objective was to build a network intrusion detector—a prediction model that can distinguish between "good" connections and "bad" connections, such as invasions or attacks. This competition resulted in the collection of a substantial amount of internet traffic records, which were then aggregated to form the KDD'99 data set. This led to the development of the NSL-KDD data set at the University of New Brunswick, which is a revised and improved version of the KDD '99.

\begin{figure}[ht!]
     \begin{center}
        \subfigure[]%Caption of First Figure
        {
            % \label{fig:1}
            \includegraphics[width=0.45\textwidth]{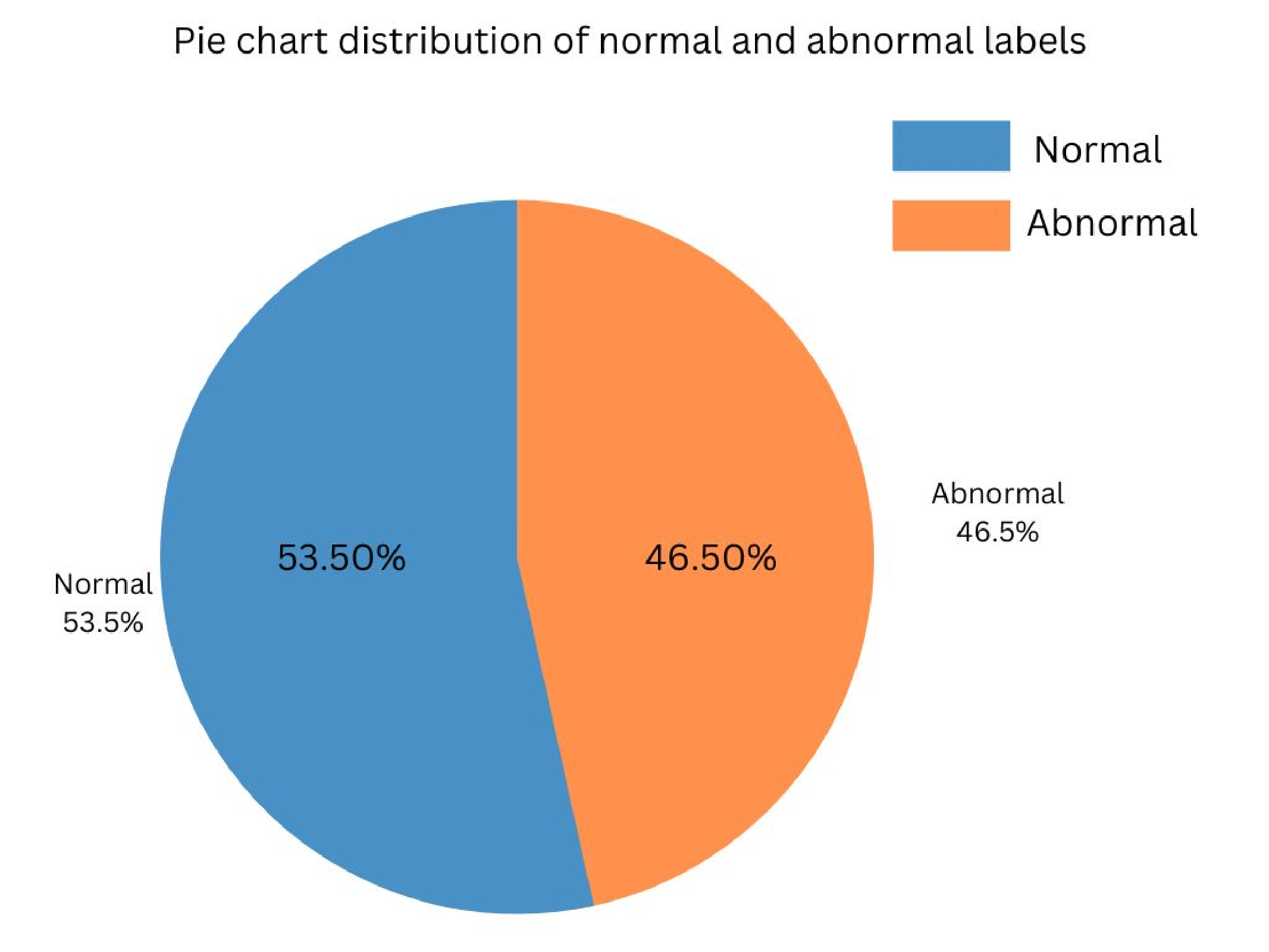}
        }%
        \subfigure[]%Caption of Second Figure
        {%
          % \label{fig:2}
          \includegraphics[width=0.45\textwidth]{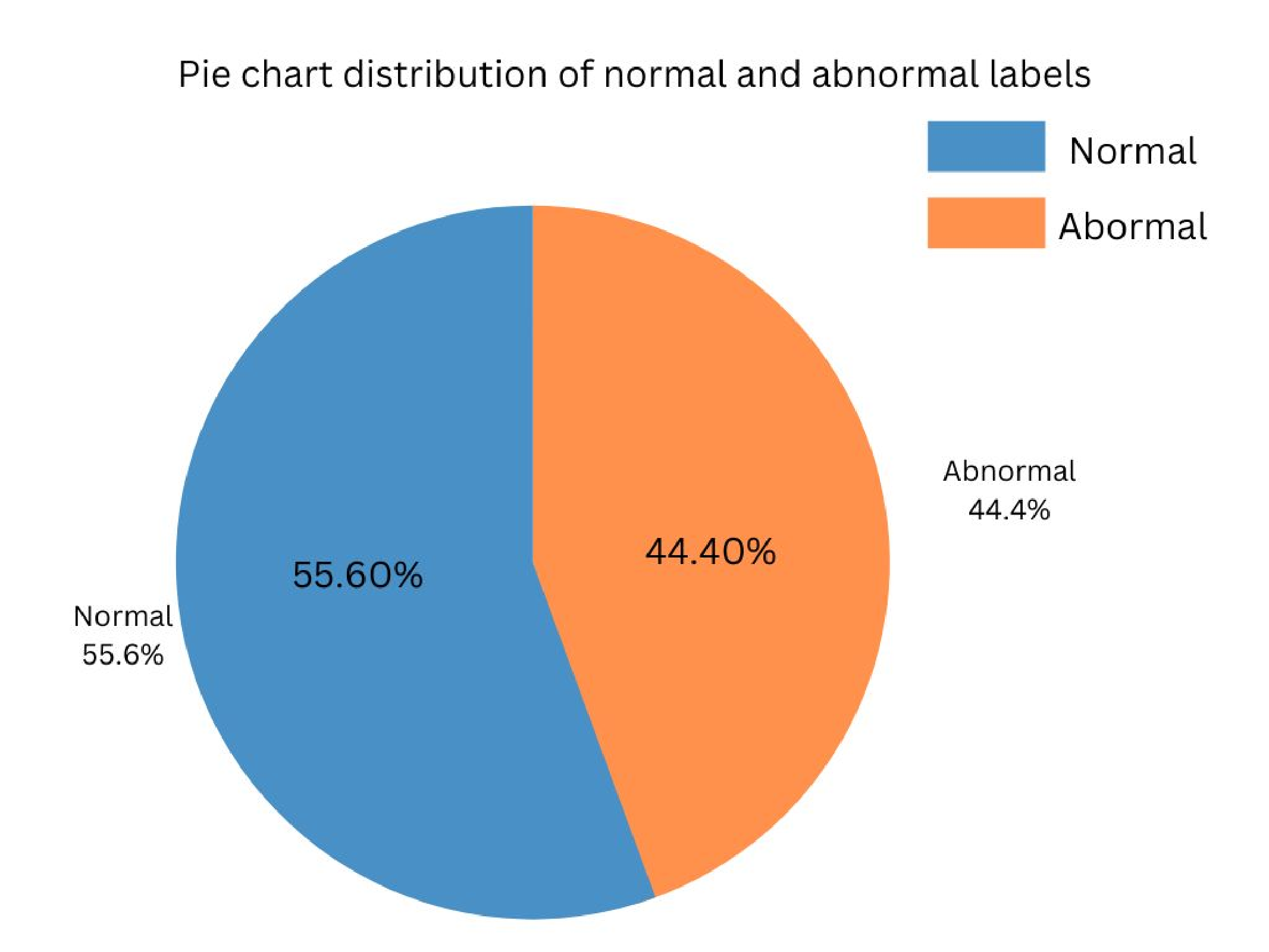}
        }\\ %  ------- End of the first row ----------------------%
        \subfigure[]%Caption of Third Figure
        {%
            % \label{fig:3}
            \includegraphics[width=0.45\textwidth]{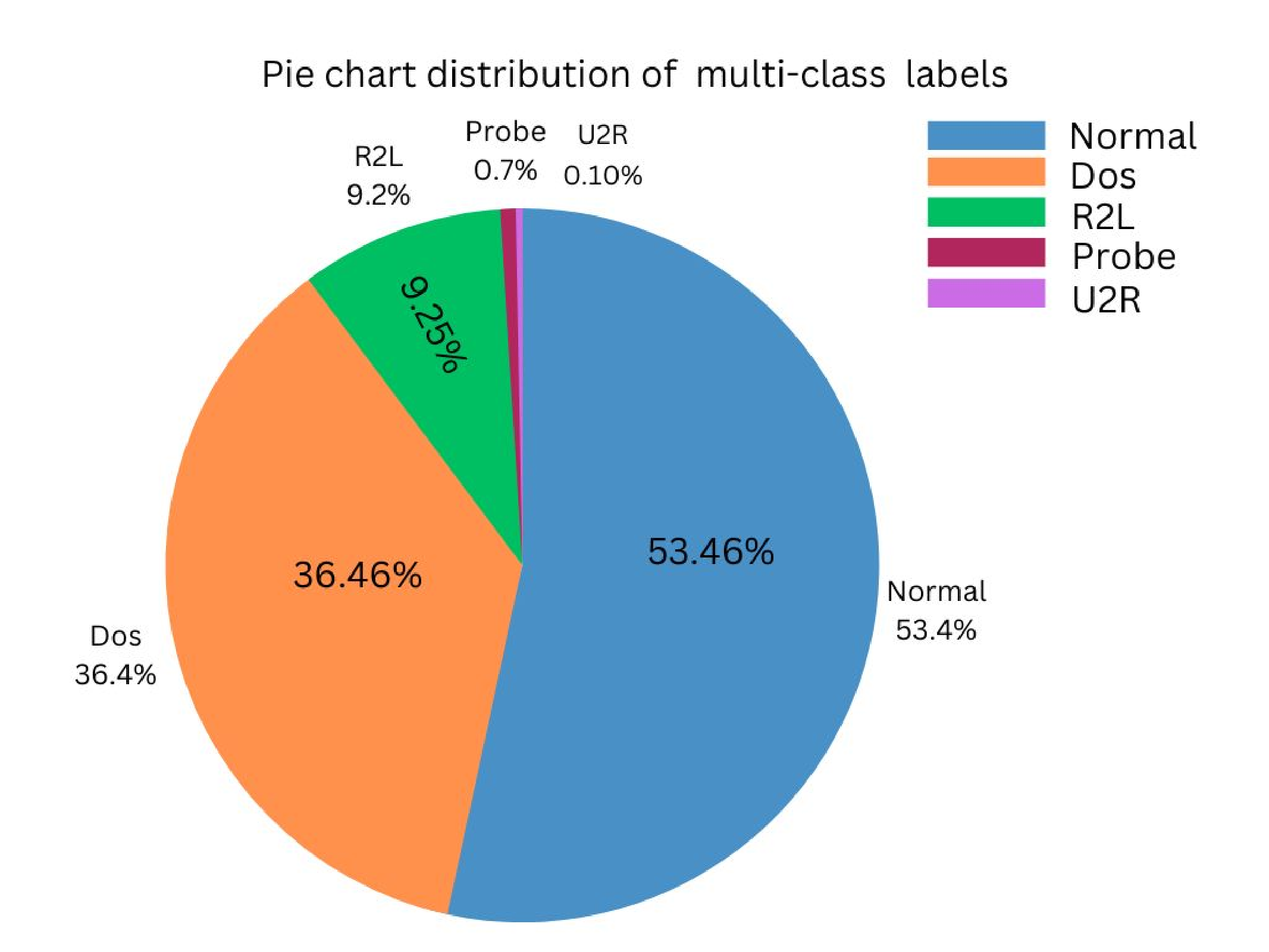}
        }
        \subfigure[]%Caption of Third Figure
        {%
            % \label{fig:4}
            \includegraphics[width=0.45\textwidth]{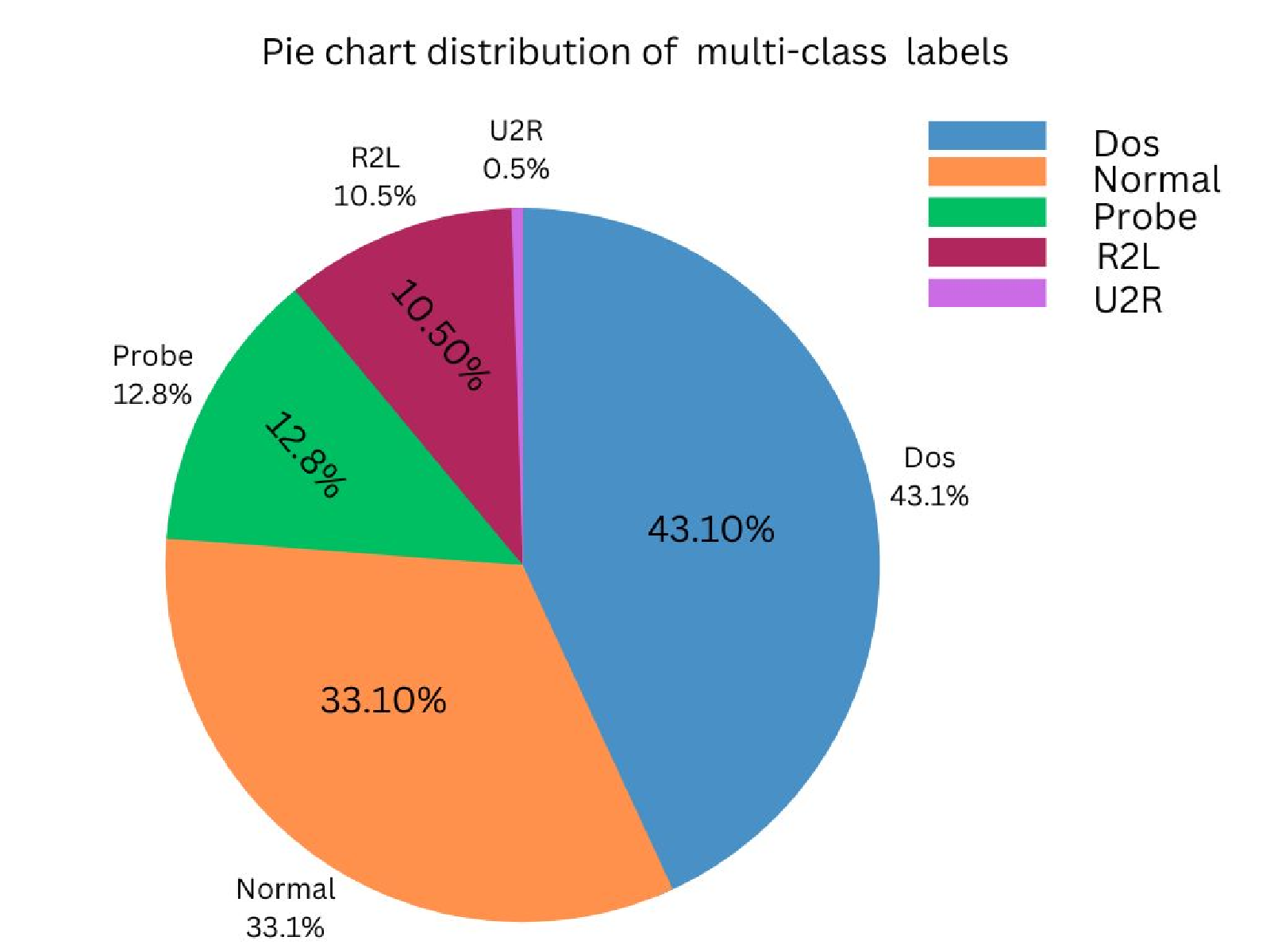}
        }
    \end{center}
    \caption{%
        \textbf (a) Distribution of Training Data for Attack Detection, \textbf (b) Distribution of Test Data for Attack Detection, \textbf (c) Distribution of Train Data for Attack Type Detection, and \textbf (d) Distribution of Test Data for Attack Type Detection}
  \label{fig:data_desc}
\end{figure}

Despite the fact that KDDTest-21 and KDDTrain+ 20 Percent are subsets of KDDTrain+ and KDDTest+, this data collection is made up of four sub-datasets: KDDTest+, KDDTest-21, KDDTrain+, and KDDTrain+ 20 Percent. KDDTrain+ will now be referred to as train, and KDDTest+ will now be known as test. Both the KDDTrain+ 20 Percent and the KDDTrain-21 are subsets of the train dataset that exclude the records with the highest traffic difficulty (Score of 21) respectively. In spite of this, the traffic records seen in the KDDTrain+ 20 Percent and KDDTest-21 datasets are not brand-new records held outside of either dataset. Rather, they are records that already exist in the test and train, respectively.
Here we use the KDD-99 data set for two types of classification one is Binary Classification and another one is Multiclass Classification. Below is a quick explanation of each classification (Fig. \ref{fig:data_desc}):
\begin{itemize}
  \item Binary Classification: In machine learning, binary classification uses supervised learning to categorize new observations into one of two classes.
  \item  Multiclass Classification: The challenge of classifying cases into one of three or more categories in statistical classification and machine learning is known as multiclass classification, also referred to as multinomial classification.
\end{itemize}

\subsection{Data Preprocessing}
First, we determine how many attack labels are present using the subclass columns. All assault labels are then classified according to the appropriate attack class. The DOS, R2l, Probe, and U2R attack classes are the four available. The attack class is then distributed, and we count every value from each category. Following that, we discover two different categories of variables: category and numerical variables. The remaining columns in this table represent the numerical variables from this dataset, with just three columns representing the category variables. In this case, we prepared the data using two distinct processes before fitting it to a specific ML model. These procedures include "On-hot-encoding" for the preparation of categorical variables and "Normalization" for the preparation of numerical variables.

\subsubsection{One hot Encoding for categorical variables}
The majority of machine learning tutorials and tools demand that you prepare your data before fitting it to a specific ML model. Changing categorical data variables so that machine learning algorithms may use them to create better predictions is one popular encoding strategy. One hot encoding is a crucial part of feature engineering for machine learning. To improve predictions and get the data ready for an algorithm, data can be changed via one-hot encoding. For each categorical value, we construct a new category column using one-hot and assign it a binary value of 1 or 0. Each integer value is represented by a binary vector. All values for the index, which is represented by a 1, are zero.

One-hot encoding has advantages for data that are unrelated to one another. Machine learning algorithms consider the organization of the numbers to be an important attribute. Or, to put it another way, individuals will consider a larger number to be more important or superior than one that is lower. However, some input data lacks ordering for category values, which can lead to inaccurate predictions and poor performance even though this is advantageous in some ordinal situations. Then one hot encoder comes to the rescue. Thanks to one-hot encoding, our training data is easier to scale and more valuable and expressive. Using numerical values allows us to more rapidly determine a probability for our values.

\subsubsection{Data Normalization for Numerical Variables}
Here, we normalize our dataset using the Standardization scaling approach. Standardization The practice of centering data around the mean while adding a unit standard deviation is known as scaling, which is frequently referred to as Z-score normalization. The property is therefore set to zero, and the resulting distribution has a unit standard deviation. By quantitatively subtracting the feature value from the mean and dividing the result by the standard deviation, standardization can be calculated. Standardization (Eqn. 1) can therefore be stated as follows:

% \begin{equation} \label{eq}
% \begin{split}
%   % X^{'} & =  \frac{X-\mu}{\sigma}
%   X^{'} = \frac{X - \mu}{\sigma}
% \end{split}
% \end{equation}

\begin{eqnarray}\label{eq}
% X' &=& \frac{X - \mu}{\sigma} \\
Y' &=& \frac{Y - \nu}{\tau}
\end{eqnarray}

In this case, it stands for the feature values' standard deviation and for their mean.
In this case, it stands for the feature values’ standard deviation and for their mean. The standardization technique does not, however, limit feature values to a particular range as the Min-Max scaling technique does. This method is beneficial for different distance-based machine learning methods, including KNN, K-means clustering, Principal component analysis, etc. It’s also crucial that the model is founded on presumptions and that the data is normally distributed.

\subsection{Class Imbalance Issue (For Multiclass)}
A significant machine learning difficulty is classification issues. There are many difficulties we must overcome in the data when we attempt to categorize a dataset based on an input dataset. One such difficulty is a dataset that is unbalanced. It has two classes, one of which is far higher than the other. There are numerous methods for dealing with it, and we employ SMOTE \cite{18_chawla2002smote} the Synthetic Minority Oversampling Technique for our dataset.

\begin{figure}
	\centering
	\includegraphics[ width=0.85\textwidth, height=0.5\textwidth]{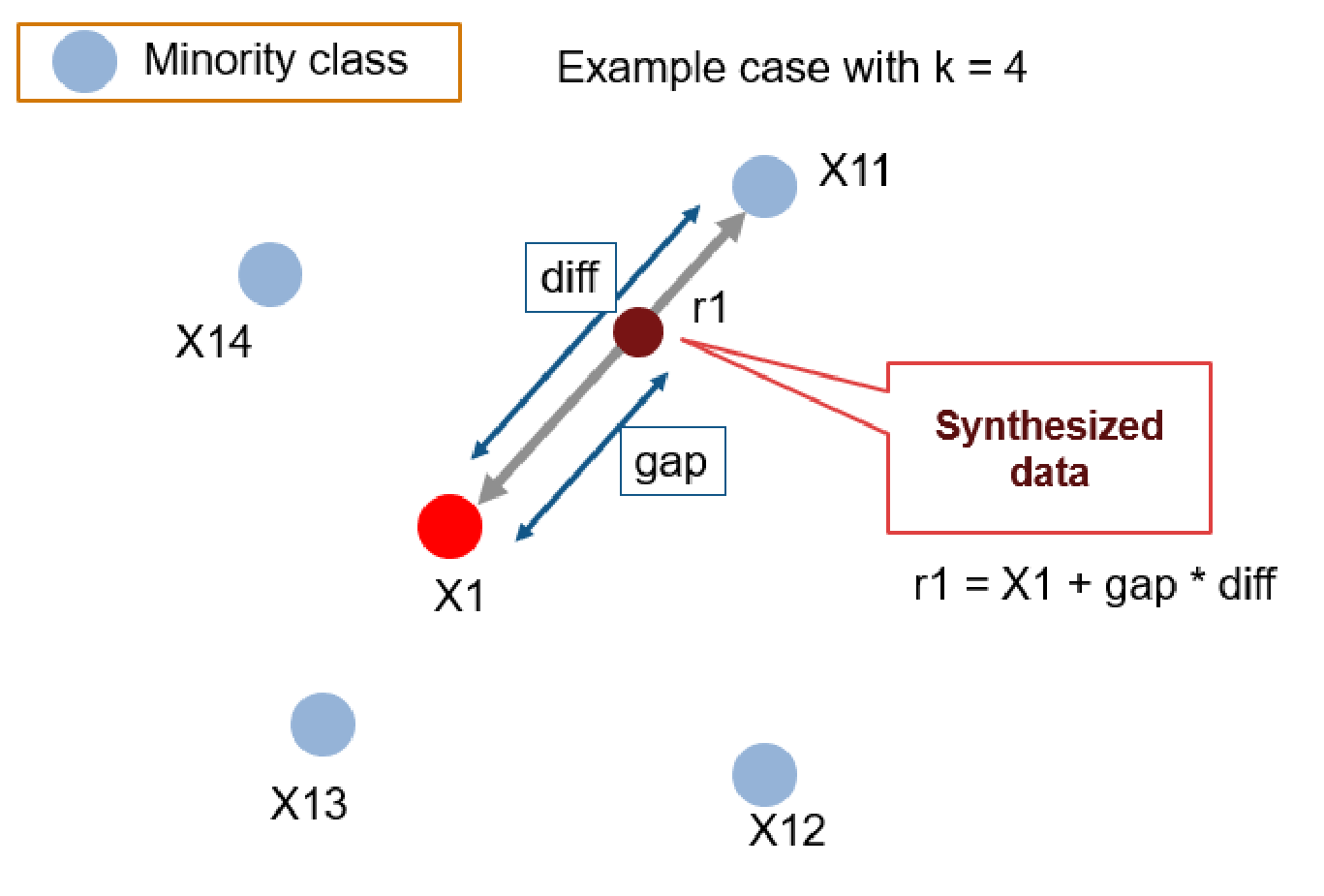}
	\caption{SMOTE the Synthetic Minority Oversampling Technique \cite{18_chawla2002smote}.}
	\label{fig:SMOTE}
\end{figure}

Considering that SMOTE is a technique for oversampling that provides artificial samples just for the minority class. This strategy aids in addressing the overfitting issue brought on by random oversampling. By interpolating between positively correlated examples that reside close together, it concentrates on the feature space to create new instances  (Fig. \ref{fig:SMOTE}).

\subsection{Data Split and Model Training}
As previously mentioned, various train datasets and test datasets from our dataset are used for model training and testing. In order to prevent data overfitting, we employed the k-fold cross-validation method here.

\begin{figure}
	\centering
	\includegraphics[ width=0.95\textwidth, height=0.60\textwidth]{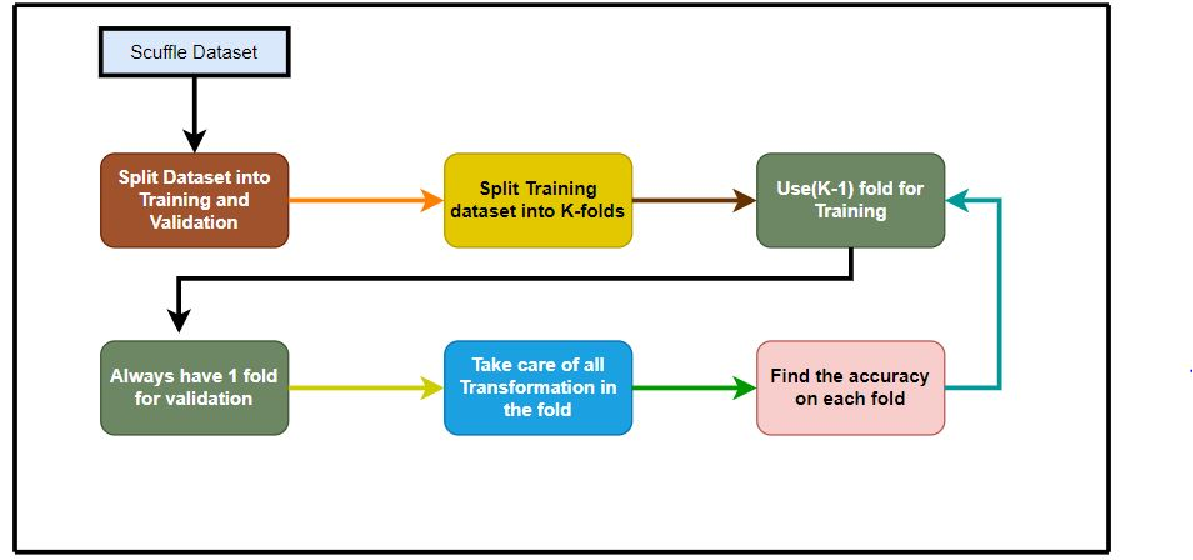}
	\caption{Step by Step Flow of K-Fold Cross-Validation.}
	\label{fig:kf}
\end{figure}

\begin{figure}
	\centering
	\includegraphics[ width=0.95\textwidth, height=0.60\textwidth]{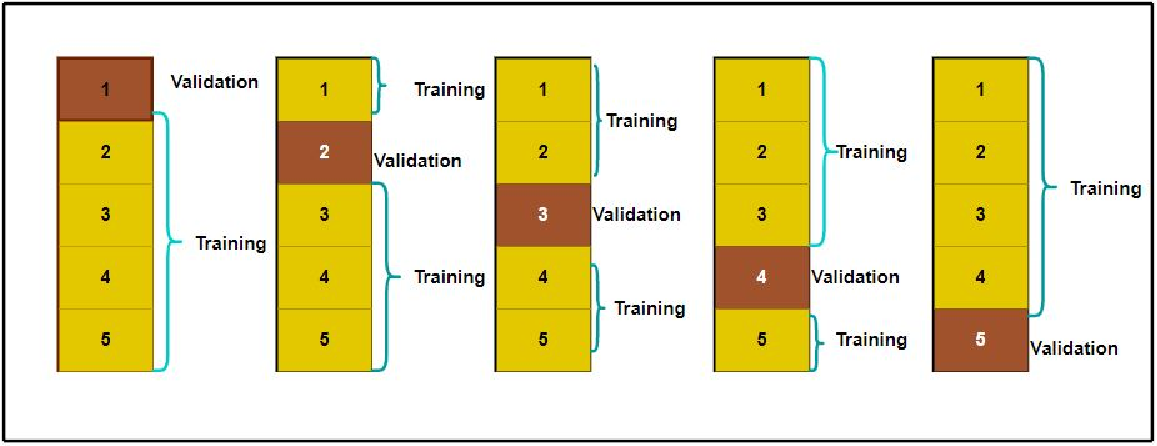}
	\caption{Cross Validation Process.}
	\label{fig:kf2}
\end{figure}

Training and testing would be carried out precisely once for each set (fold) throughout the entire process. It helps to avoid over fitting. We are aware that training a model with all of the data in a single, quick run yields the highest performance accuracy. By avoiding this k-fold cross-validation, we can create a generalized model  (Fig. \ref{fig:kf} and Fig. \ref{fig:kf2}). Here, the Test and Train data set will assist in evaluations of the building model and its hyperparameters.

The data collection does not need to be divided because we already have various datasets for testing and training. Here, training data have been divided into separate sets for training and validation. For our cross-validation value for K=5,  we are partitioning the given dataset into 5 folds and doing the training and validation. One fold from each run is taken into account for validation, with the remaining folds used for training and continuing with iterations. Following that, test datasets are subjected to the same data pre-processing procedures as training datasets in order to acquire curated forms before testing can begin.

\subsection{Machine Learning Models}
We used Logistic Regression, Decision Tree Classifier, Random Forest Classifier, Multinomial NB Classifier, SVM with Linear kernel and SVM with RBF kernel, Extreme Gradient Boosting and Adaptive Boosting for classification. Besides that, we have also developed our own deep learning architecture SCGNet and defined the architecture with Random Search and estimated the hyperparameters with Hyperband.

\begin{itemize}
  \item \textbf{Multinomial NB Classifier: }Multinomial Naive Bayes is a popular supervised learning classifier for categorical text analysis, based on Bayes theorem. It guesses text tags and outputs the highest probability tag for a given sample.
  
  \item \textbf{SVM with Kernel: }The kernels are a set of mathematical operations used by SVM algorithms. A kernel's job is to take data as input and change it into the required form. The scalar product between two points in an incredibly appropriate feature space is returned by the kernel functions. Thus, even in the situation of very high-dimensional spaces, by defining a notion of similarity, with little computational expense. 
  
  \item \textbf{Extreme Gradient Boosting: }XGBoost improves model performance and execution speed, allowing for larger datasets and outperforming current models. In terms of model performance and execution speed, it outperforms other methods like RF, GBM, and GBDT.

  \item \textbf{AdaBoost: }AdaBoost, also known as Adaptive Boosting, is a predictive modelling algorithm used as an ensemble. It uses decision trees with one level or split as estimators. AdaBoost assigns equal weights to data pieces, assigning larger weights to incorrectly classified points, and training models until a smaller error is observed. 

\end{itemize}

% \subsubsection{Multinomial NB Classifier}
% Multinomial Naive Bayes is a popular supervised learning classifier for categorical text analysis, based on Bayes theorem. It guesses text tags and outputs the highest probability tag for a given sample. 

% \subsubsection{SVM with Kernel}
% The kernels are a set of mathematical operations used by SVM algorithms. A kernel's job is to take data as input and change it into the required form. The scalar product between two points in an incredibly appropriate feature space is returned by the kernel functions. Thus, even in the situation of very high-dimensional spaces, by defining a notion of similarity, with little computational expense. 

% \subsubsection{Extreme Gradient Boosting}
% XGBoost improves model performance and execution speed, allowing for larger datasets and outperforming current models. In terms of model performance and execution speed, it outperforms other methods like RF, GBM, and GBDT.

% \subsubsection{AdaBoost}
% AdaBoost, also known as Adaptive Boosting, is a predictive modelling algorithm used as an ensemble. It uses decision trees with one level or split as estimators. AdaBoost assigns equal weights to data pieces, assigning larger weights to incorrectly classified points, and training models until a smaller error is observed. 

\subsection{Deep Learning Architecture}
We have developed a Hybrid Deep Learning Architecture by stacking 1D Convolution and GRU. We have experimented with the depth, number of nodes, dropout percentage and activation functions. These parameters are defined using RandomSearch and HyperBand. Following table shows the hyperparameters (Table. \ref{tab: hyper}). Now we describe the final architecture of the Model:
\begin{itemize}
  \item Finally we have used 2 Convolution Block and 2 GRU Block.
  \item First Convolution Block contains 1 Conv1d layer with 32 kernels of size 2 followed by Batch Normalization and MaxPooling of kernel size 2 with dropout 0.2.
  \item Second Convolution Block contains 1 Conv1d layer with 64 kernels of size 2 followed by Batch Normalization and MaxPoolin of the kernel size 2 with dropout 0.2.
  \item First GRU Block contains 100 GRU Nodes with Batch Normalization and Dropout 0.2.
  \item Second GRU Block contains 100 GRU Nodes with Batch Normalization and Dropout 0.2.
  \item Then they are flattened and a Dense layer of 64 nodes followed by a Dropout of 0.5.
  \item For binary, we have 1 classification node with a sigmoid function, and for multiclass we have 5 nodes with a softmax activation function.
  \item All the layers have Relu as an activation function.

\end{itemize}

\begin{table}[!htbp] \centering
\caption{Hyper Parameter List}
\label{tab: hyper}
\begin{tabular}{@{\extracolsep{5pt}} llrrr} 
\\[-1.8ex]\hline 
\hline \\[-1.8ex] 
\multicolumn{1}{c}{Name of Hyper Parameter} & \multicolumn{1}{c}{Values} \\
\hline \\[-1.8ex] 
\midrule
  	Number of Kernels of Conv1d & 32, 64, 128\\
  	Kernel Size for Conv1d & 2, 5, 7\\
  	Dropout percentage & 0.2, 0.4, 0.5\\
        Number of Convolution Block & 1, 2, 3, 4\\
  	Number of GRU Node & 100, 200, 300\\
  	Number of GRU Block & 1, 2, 3, 4\\
  	Activation Functions & Relu, elu, Relu-6\\

\hline \\[-1.8ex] 
\end{tabular}
\end{table}

\section{Result Analysis}
\label{sec_result_analysis}
We have used four metrics to assess our classification models: accuracy, 11-score, precision, and recall \cite{19_b_2020}. The experiment was implemented using Python library packages like keras, numpy, sci-kit learn, pandas, and matplotlib. The experiment was run and trained in NVIDIA 3060 GPU. The models were trained for 500 epochs with batch size 256 and the learning rate was set to 0.01 with exponential decay. Adam optimizer was used. Kears Autotuner was used for hyperparameter search and had a callback function that monitors validation loss and accuracy. Models were early stopped using early stopping having a patience value for 15 epochs.

All the results are generated on the test dataset provided in the NSL\_KDD dataset. We can see that for both tasks, our model outperforms the current SOTA in terms of accuracy and other metrics. Results are shown in Table. \ref{tab:result_for_b_inary_classification} and Table. \ref{tab:result_for_b_inary_classification2}.

\begin{table}[!htbp] \centering
\caption{Result for Binary Classification.}
\label{tab:result_for_b_inary_classification}
\begin{tabular}{@{\extracolsep{5pt}} llrrr} 
\\[-1.8ex]\hline 
\hline \\[-1.8ex] 
\multicolumn{1}{c}{Algorithm } & \multicolumn{1}{c}{Accuracy} & \multicolumn{1}{c}{Precision } & \multicolumn{1}{c}{Recall} & \multicolumn{1}{c}{F1-Score} \\
\hline \\[-1.8ex] 
\midrule
 	LR & 0.548 & 0.56 & 0.96 & 0.721 \\
  	DT & 0.605 & 0.79 & 0.41 & 0.863 \\
  	RF & 0.751 &  0.94 & 0.60 & 0.63 \\
  	SVM & 0.792 &  0.86 & 0.58 & 0.791\\
  	NB & 0.525 & 0.55 & 0.90 & 0.762 \\
        KNN & 0.796 & 0.801 & 0.796 & 0.798 \\
  	XGBoost & 0.813 & 0.814 & 0.813 & 0.836 \\  
        AdaBoost & 0.856 & 0.826 & 0.896 & 0.866 \\
        SCGNet & 0.996 & 0.991 & 0.993 & 0.992 \\
\hline \\[-1.8ex] 
\end{tabular}
\end{table}

\begin{table}[!htbp] \centering
\caption{Result for Multiclass Classification.}
\label{tab:result_for_b_inary_classification2}
\begin{tabular}{@{\extracolsep{5pt}} llrrr} 
\\[-1.8ex]\hline 
\hline \\[-1.8ex] 
\multicolumn{1}{c}{Algorithm } & \multicolumn{1}{c}{Accuracy} & \multicolumn{1}{c}{Precision } & \multicolumn{1}{c}{Recall} & \multicolumn{1}{c}{F1-Score} \\
\hline \\[-1.8ex] 
\midrule
 	LR & 80.20 & 81.44 & 80.20 & 76.22 \\
  	DT & 76.80 & 75.70 & 76.80 & 76.10 \\
  	RF & 78.93 & 78.28 & 78.93 & 75.18 \\
  	MNB & 82.60 &  81.74 & 82.60 & 81.56\\
  	KNN & 80.48 & 80.69 & 80.48 & 80.58 \\
        Linear SVM & 76.80 & 79.51 & 76.80 & 69.60 \\
  	RBF SVM & 78.36 & 80.39 & 78.36 & 72.78 \\  	
        XGBoost & 76.94 & 76.94 & 76.94 & 70.85 \\
        SCGNet & 99.50 & 89.90 & 92.30 & 91.20 \\

\hline \\[-1.8ex] 
\end{tabular}
\end{table}

\section{Conclusion and Future Work}
\label{sec_conclusion_and_future_work}
In this paper based on our examination of neural networks and intrusion detection systems, we developed a new architecture and our results were compared against various SOTA models. We compare the performance considering some features, accuracy, precision, recall and f1-score. 

We want to experiment with our data pipeline and model on SoTA datasets that are available like UNSW-NB1, CSE-CIC-IDS2018, and others. We also want to reduce our model size by applying pruning and quantization. Another important part we would like to address is feature explainability. Thus it will be a complete tool for industrial use cases.

\let\cleardoublepage\clearpage

%\bibliographystyle{plain}
%\bibliography{bibtex_example}
%\bibliography{references}